\documentclass[conference]{IEEEtran}
\IEEEoverridecommandlockouts

\usepackage{cite}
\usepackage{amsmath,amssymb,amsfonts}
\usepackage{bm}
\usepackage{algorithm}
\usepackage{algpseudocode}
\usepackage[hidelinks]{hyperref}
\usepackage{indentfirst}
\usepackage{amsthm}
\usepackage{mathtools}
\usepackage{extarrows}
\usepackage{graphicx}
\usepackage{textcomp}
\usepackage{xcolor}
\usepackage{xspace}
\newcommand{\herm}{{\mathsf{H}}}
\newcommand{\trans}{{\mathsf{T}}}
\newcommand{\E}[1]{\mathbb{E}\left[ #1 \right]}
\newcommand{\irmg}{\textsc{IRM Games}\xspace}
\newcommand{\firmg}{\textsc{F-IRM Games}\xspace}
\newcommand{\virmg}{\textsc{V-IRM Games}\xspace}
\newcommand{\fedavg}{\textsc{FedAVG}\xspace}
\newcommand{\fflg}{\textsc{F-FL Games}\xspace}
\newcommand{\vflg}{\textsc{V-FL Games}\xspace}
\newcommand{\flg}{\textsc{FL Games}\xspace}
\algdef{SE}[SUBALG]{Indent}{EndIndent}{}{\algorithmicend\ }%
\algtext*{Indent}
\algtext*{EndIndent}
\usepackage{subcaption}

\usepackage{soul}

\definecolor{darkorange}{rgb}{1.0, 0.55, 0.0}

\begin{document}

\title{Federated Learning Games for Reconfigurable Intelligent Surfaces via Causal Representations}

\author{\IEEEauthorblockN{%
Charbel Bou Chaaya, Sumudu Samarakoon, and Mehdi Bennis
}
\IEEEauthorblockA{%
Centre for Wireless Communications, University of Oulu, Finland\\
Emails: \{charbel.bouchaaya, sumudu.samarakoon, mehdi.bennis\}@oulu.fi}
}

\maketitle

\begin{abstract}
In this paper, we investigate the problem of robust Reconfigurable Intelligent Surface (RIS) phase-shifts configuration over heterogeneous communication environments. 
The problem is formulated as a distributed learning problem over different environments in a Federated Learning (FL) setting. 
Equivalently, this corresponds to a game played between multiple RISs, as learning agents, in heterogeneous environments.
Using Invariant Risk Minimization (IRM) and its FL equivalent, dubbed FL Games, we solve the RIS configuration problem by learning invariant causal representations across multiple environments and then predicting the phases. 
The solution corresponds to playing according to Best Response Dynamics (BRD) which yields the Nash Equilibrium of the FL game. 
The representation learner and the phase predictor are modeled by two neural networks, and their performance is validated via simulations against other benchmarks from the literature. 
Our results show that causality-based learning yields a predictor that is 15\% more accurate in unseen Out-of-Distribution (OoD) environments.
\end{abstract}


\section{Introduction}
The advent of Reconfigurable Intelligent Surfaces (RISs) will substantially boost the performance of wireless communication systems. 
These surfaces are manufactured by layering stacks of sheets made out of engineered materials, called meta-materials, built on a planar structure. The reflection coefficients of the meta-material elements, called meta-atoms, vary depending on their physical states. Thus, the direction of incident electromagnetic waves on RISs can be manipulated with the aid of simple integrated circuit controllers that modify meta-atoms' states. 
In this view, the RIS technology provides a partial control over the wireless propagation environment rendering improved spectral efficiency with a minimal power footprint \cite{huang2019reconfigurable}. RIS is considered a fundamental enabler to achieve the 6G vision of smart radio environments \cite{renzo2019smart}.

One of the major challenges in the RIS technology is the accurate tuning of RIS phases.
To this extent, a vast majority of the existing literature on RIS-assisted communication relies on the use of Channel State Information (CSI) to train Machine Learning (ML) models that predict the optimal RIS configuration \cite{ozdougan2020deep, stylianopoulos2022deep, alexandropoulos2022pervasive}. 
These methods seek either a centralized-controller driven approach, or a distributed multi-agent optimization technique, such as Federated Learning (FL). Other works such as \cite{park2022extreme, alexandropoulos2020phase} exploit the users' locations to employ a location-based passive RIS beamforming. Either way, their main focus is to draw on the statistical correlations of the observed data, while overlooking the impacts of heterogeneous system designs (e.g., different RISs, propagation environments, users distribution, etc.). Moreover, these approaches produce a high inference accuracy within a fixed environment, from which the training and testing data are drawn. They subsequently fail to have a good Out-of-Distribution (OoD) generalization in unseen environments.

Although FL provides a learning framework where multiple agents train a collaborative model while preserving privacy, its state-of-the-art approach, such as Federated Averaging (\fedavg) \cite{mcmahan2017communication}, is known to perform poorly when the local data is non-identical across participating agents. This is due to the fact that \fedavg (and its variants) solve the distributed learning problem via Empirical Risk Minimization (ERM), that minimizes the empirical distribution of the local losses assuming that the data is identically distributed. 
To mitigate this issue, the authors in \cite{issaid2021federated} leveraged Distributionally Robust Optimization (DRO) \cite{deng2020distributionally} and proposed a federated DRO for RIS configurations. Therein, the distributed learning problem is cast as a minimax problem, where the model's parameters are updated to minimize the maximal weighted combination of local losses. This ensures a good performance for the aggregated model over the worst-case combination of local distributions.

On the other hand, \cite{9593252} used Invariant Risk Minimization (IRM) \cite{arjovsky2019invariant} to formulate the problem of learning optimal RIS phase-shifts. The aim of IRM is to capture causal dependencies in the data, that are invariant across different environments. In \cite{9593252}, the authors empirically showed that using relative distances and angles between the RIS and the transmitter/receiver as causal representations for the CSI, improves the robustness of the RIS phase predictor.
However, these representations were not learned by the configuration predictor, but were predefined and fixed. Also, this solution assumes that multiple environments are known to the predictor beforehand, which is an unfeasible assumption.

The main contribution of this paper is a novel distributed IRM-based solution to the RIS configuration problem.
We cast the problem of RIS phase control as a federated learning problem with multiple RISs controllers defined over heterogeneous environments, using a game-theoretic reformulation of IRM, referred to as Federated Learning Games (\flg) \cite{ahuja2020invariant, gupta2022fl}. The solution of this problem is proven to be the Nash Equilibrium of a strategic game where each RIS updates its configuration predictor by minimizing its local loss function. This game is indexed by a representation learner that is shared among the RISs to extract a causal representation from the CSI data. The representation learner and predictors are trained in a distributed and supervised learning manner.
The numerical validations yield that the proposed design improves the accuracy of the predictor tested in OoD settings by 15\% compared to state-of-the-art RIS designs.

The remainder of this paper is organized as follows. In Section \ref{section:2}, we describe the system model and conventional approaches to the RIS configuration problem using \fedavg. The \flg solution that involves extracting causal representations from the data is discussed in Section \ref{section:3}. 
Section \ref{section:4} presents the simulation results that compare the proposed algorithm with benchmarks. Concluding remarks are drawn in Section \ref{section:5}.

\section{System Model and Problem Formulation} \label{section:2}
\begin{figure}
    \centering
    \includegraphics[width=\linewidth]{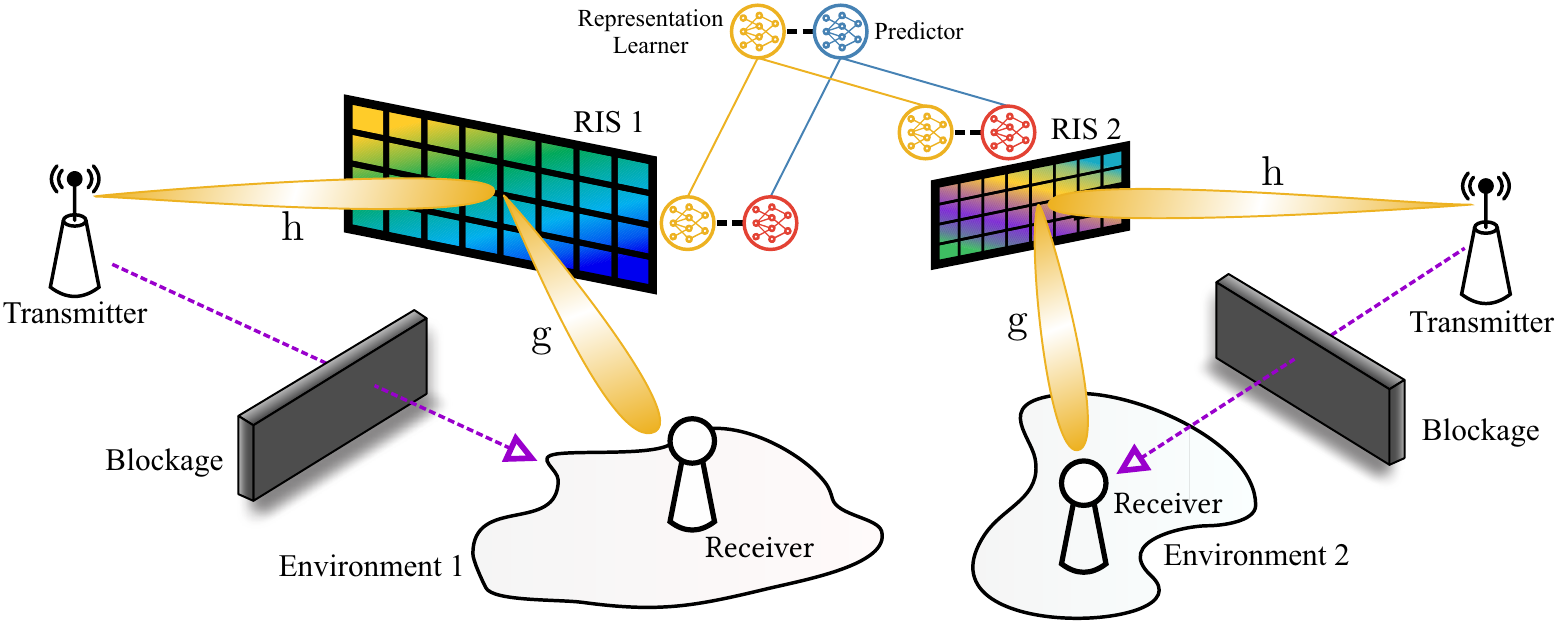}
    \caption{System model illustrating different RIS-assisted communication scenarios. Each RIS is conceived differently from other RISs, and serves differently distributed receivers.}
    \label{fig:system_model}
\end{figure}
We consider a set of environments $\mathcal{R}$ where each environment $r\in\mathcal{R}$ consists of a RIS-assisted downlink communication between a transmitter (Tx) and a receiver (Rx) as shown in Fig. \ref{fig:system_model}.
Both the Tx and the Rx are equipped with a single antenna each, and we assume that the direct link between them is blocked in which, the channel of direct link is dominated by the reflected channel.
The RIS in environment $r$ is composed by $N^r=N^r_\mathrm{x} N^r_\mathrm{y}$ reflective elements where $N_\mathrm{x}^r$ and $N_\mathrm{y}^r$ are the number of horizontal and vertical reflective elements, respectively.
Additionally, the inter-element distances over horizontal and vertical axes are characterized by $d_\mathrm{x}^r$ and $d_\mathrm{y}^r$.
Each RIS element applies a phase shift on its incident signal and the reflected signals are aggregated at the Rx. 
Note that the location of the Tx is fixed while the location of Rx is arbitrary, which is sampled by a predefined distribution. 
The choices of the Rx distribution along with the parameters $(N_\mathrm{x}^r,N_\mathrm{y}^r,d_\mathrm{x}^r,d_\mathrm{y}^r)$ collectively define the environment $r$.
We assume that these environments are completely separate, i.e. each Rx receives only the signal reflected by the RIS in its corresponding environment.

\subsection{Channel Model}
For the notation simplicity, we have omitted the notion of environment during the discussion within this subsection.
Let $\mathbf{g} \in \mathbb{C}^{N}$ be the channel between the RIS and the Rx, which is dominated by its line-of-sight (LoS) component. 
Hence, by denoting $\varphi_r$ and $\vartheta_r$ as the azimuth and elevation angles-of-departure (AoD) from the RIS respectively, the channel is modeled as
\begin{equation}\label{eq:channelg}
    \mathbf{g} = \sqrt{\alpha_r} \; \mathbf{a}_N\left(\varphi_r, \vartheta_r\right),
\end{equation}
where $\alpha_r$ represents the path-loss.
Additionally, we define the steering function:
\begin{equation}
    \mathbf{a}_N\left(\varphi, \vartheta\right) = \left[ e^{ \tfrac{2\pi j}{\lambda}  \Delta_1\left( \varphi,\vartheta\right)  }, \cdots, e^{  \tfrac{2\pi j}{\lambda}  \Delta_N\left( \varphi,\vartheta\right) } \right]^\trans,
\end{equation}
and a set of operators for $n=1,\dots,N$:
\begin{align}
    \Delta_n\left(\varphi,\vartheta\right) ={i_N(n) d_\mathrm{x} \cos(\vartheta) \sin(\varphi) +  j_N(n) d_\mathrm{y} \sin(\vartheta)}, \\
    i_N(n) = \left(n-1\right)  \;  \mathrm{mod} \;  N_\mathrm{x},
    \qquad
    j_N(n) = \left\lfloor \frac{n-1}{N_\mathrm{x}} \right\rfloor,\label{eq:indices}
\end{align}
where $\lambda$ is the wavelength, $\mathrm{mod}$ and $\lfloor\cdot\rfloor$ denote the modulus and floor operators. On the other hand, the channel $\mathbf{h} \in \mathbb{C}^{N}$ between the Tx and the RIS will have both LoS and non line-of-sight (NLoS) components. Therefore, we model $\mathbf{h}$ using Rician fading with spatial correlation, since the RIS elements are closely distanced. Accordingly, we have:
\begin{equation}\label{eq:channelh}
    \mathbf{h} = \sqrt{\alpha_t} \,
    \left(\sqrt{\frac{\kappa}{1+\kappa}} \, \overline{\mathbf{h}}
    + \sqrt{\frac{1}{1+\kappa}} \, \widetilde{\mathbf{h}}\right),
\end{equation}
where $\alpha_t$ and $\kappa$ are the path-loss and the Rician coefficient respectively, $\overline{\mathbf{h}}$ is the LoS factor, and $\widetilde{\mathbf{h}}$ represents the small scale fading process in the NLoS component. Further, for the LoS link, the LoS factor is:
\begin{equation}
    \overline{\mathbf{h}} = \mathbf{a}_N\left(\varphi_t, \vartheta_t\right),
\end{equation}
where $\left(\varphi_t, \vartheta_t\right)$ are the angles-of-arrival (AoA) to the RIS. We model the NLoS link as $\widetilde{\mathbf{h}} \sim \mathcal{CN}\left(\mathbf{0}_N, \mathbf{R}\right)$, where $\mathbf{R}$ is a covariance matrix that captures the spatial correlation among the channels of the RIS elements. In the case of isotropic scattering in front of the RIS, a closed-form expression for $\mathbf{R}$ is \cite[Proposition 1]{bjornson2020rayleigh}:
\begin{equation}
    \left[\mathbf{R}\right]_{m,n} = \mathrm{sinc}\left(\frac{2 \, \lvert\mathbf{u}_m - \mathbf{u}_n\rvert}{\lambda}\right) \qquad m, n = 1,\dots,N,
\end{equation}
where $\mathbf{u}_n = \left[i_N(n) \, d_\mathrm{x}, j_N(n) \, d_\mathrm{y}\right]^\trans$ represents the locations of the $n$\textsuperscript{th} element with $i_N$ and $j_N$ being the horizontal and vertical indices given in \eqref{eq:indices}, and $\mathrm{sinc}(\cdot)$ is the normalized sampling function.

\subsection{Downlink Rate Maximization}
At every transmission slot, in each environment $r \in \mathcal{R}$, the RIS selects its phases in order to enhance the downlink rate at the Rx. 
Let $\bm{\theta}^r = \left[\theta_1^r,\dots,\theta_N^r\right]^\trans$ denote the phase decisions at the RIS. Thus, the received signal at the Rx is:
\begin{equation}
    y_r = \left(\mathbf{h}_r^{\herm} \bm{\Theta}_r \, \mathbf{g}_r\right)s_r + z_r,
\end{equation}
where $\bm{\Theta}_r = \mathrm{diag}\left(e^{j\theta_1^r}, \dots, e^{j\theta_N^r}\right)$ is the RIS reflection matrix, $s_r$ is the transmitted signal satisfying the power budget constraint $\E{\lvert s_r \rvert^2} = p$, and $z_r \sim \mathcal{CN}\left(0, \sigma^2\right)$ is the additive noise with power $\sigma^2$. 
In this view, the objective of downlink rate maximization can be cast as follows:
\begin{equation}
    \label{eq:problem}
    \begin{aligned}
       &\underset{\bm{\theta}^r \, \in \, \mathcal{C}}{\textrm{maximize}} \quad && 
        \log_2\left(1+\frac{\lvert\mathbf{h}_r^\herm \bm{\Theta}_r \, \mathbf{g}_r\rvert^2 p}{\sigma^2}\right),
    \end{aligned}
\end{equation}
where $\mathcal{C}$ is the set of feasible RIS configurations.

In order to solve \eqref{eq:problem}, a perfect knowledge of both channels $\mathbf{h}$ and $\mathbf{g}$ is assumed. Even under perfect CSI, determining the optimal set of phase shifts $\bm{\theta}^{r\star}$ requires a heuristic search due to the notion of configuration classes $\mathcal{C}$. Such solutions cannot be practically adopted since they are not scalable with the number of RIS elements. As a remedy, we resort to ML in order to devise a data-driven solution.

In this context, consider that the RIS in environment $r \in \mathcal{R}$ (later referred to as agent $r$) has a dataset $\mathcal{D}_r = \left\{\left(\bm{x}_j^r, c_j^r\right) \;\middle\vert\; j=1,\dots,D_r\right\}$ containing $D_r$ samples of observed CSI $\bm{x}_j^r = \left(\mathbf{h}_j^r, \mathbf{g}_j^r\right)$ that are labeled by $c_j^r$ corresponding to the optimal RIS phase shifts $\bm{\theta}^{r\star}$ solving \eqref{eq:problem}. 
We then seek to construct a mapping function $f_{\bm{w}}(\cdot)$ parameterized by $\bm{w}$, that solves:
\begin{equation}\label{eq:erm}
    \underset{\bm{w}}{\textrm{minimize}} \qquad
    \frac{1}{\lvert\mathcal{R}\rvert} \sum_{r=1}^{\lvert\mathcal{R}\rvert}
    \frac{1}{D_r} \sum_{j=1}^{D_r} \ell\left(c_j^r, f_{\bm{w}}(\bm{x}_j^r)\right),
\end{equation}
where $\ell(\cdot\,,\cdot)$ is the loss function in terms of phase prediction. In order to optimize the model parameter $\bm{w}$ in the ERM formulation in \eqref{eq:erm}, all agents are required to share their datatsets $\mathcal{D}_r$ with a central server. Due to privacy concerns and communication constraints, \eqref{eq:erm} is recast as a FL problem by minimizing a global loss function as follows:
\begin{equation}\label{eq:fl}
    \underset{\bm{w}}{\textrm{minimize}} \qquad
    \frac{1}{\lvert\mathcal{R}\rvert} \sum_{r=1}^{\lvert\mathcal{R}\rvert}
    \ell_r\left(f_{\bm{w}}\right),
\end{equation}
where $\ell_r\left(f_{\bm{w}}\right) = \frac{1}{D_r} \sum_{j=1}^{D_r} \ell\left(c_j^r, f_{\bm{w}}(\bm{x}_j^r)\right)$ is the local loss function of agent $r$. One of the most popular approaches in FL to solve \eqref{eq:fl} is the \fedavg algorithm \cite{mcmahan2017communication}.

However, the formulation in \eqref{eq:fl} assumes that all agents in $\mathcal{R}$ have an equal impact on training the global model $\bm{w}$. 
This falls under the strong assumption of uniform and homogeneous data distribution across agents. Under data heterogeneity, drifts in the agents' local updates with respect to the aggregated model might occur, since the local optima do not necessarily coincide with the global optima. 
Thus, the obtained model suffers from the instability in convergence, and  fails to generalize to OoD samples. 
To obviate this issue, we resort to Invariant Risk Minimization (IRM) \cite{arjovsky2019invariant} and its FL variant dubbed \flg \cite{gupta2022fl}.
%\vspace{-.15cm}
\section{FL Games for Phase Optimization}\label{section:3}
%\vspace{-.15cm}
The key deficiency of using \fedavg on limited datasets distributed across agents is that its trained predictor heavily relies on statistical correlations among observations. These correlations are specious since they depend on the environment from which they were sampled. Thus, overfitting to these correlations prevents \fedavg from training a predictor that is robust in unseen environments. To overcome this issue, we turn our attention to algorithms that learn causal representations that are invariant across different agents, which improves the model's OoD generalization across many environments.

In this direction, IRM \cite{arjovsky2019invariant} jointly trains an \emph{extraction function} $f_{\bm{\phi}}$ and a \emph{predictor} $f_{\bm{w}}$ across training environments $\mathcal{R}$ in such a way that $f_{\bm{w}} \circ f_{\bm{\phi}}$ generalizes well in unseen environments $\mathcal{R}_\text{all} \supset \mathcal{R}$.
The main idea is to build a parameterized feature extractor $f_{\bm{\phi}}$ that reveals the causal representations in the samples, so as to perform causal inference by optimizing $f_{\bm{w}}$. Thus, given an extraction function, the predictor $f_{\bm{w}}$ is the one that is simultaneously optimal across all training environments $\mathcal{R}$. Formally, this boils down to solving the following problem:
\begin{equation}
    \label{eq:IRM}
    \begin{aligned}
        &\underset{\bm{\phi}, \, \bm{w}}{\textrm{minimize}}&& \quad 
        \frac{1}{\lvert\mathcal{R}\rvert} \sum_{r=1}^{\lvert\mathcal{R}\rvert}
        \ell_r\left(f_{\bm{w}} \circ f_{\bm{\phi}}\right)\\
        &\textrm{subject to}&& \quad \bm{w} \in \underset{\bm{w}^\prime}{\arg\min}\,\; \ell_r\left(f_{\bm{w}^\prime} \circ f_{\bm{\phi}}\right) \;\;\;\;\;\forall\, r \in \mathcal{R}.
    \end{aligned}
\end{equation}
Note that IRM is formulated as a single agent optimization problem, and assumes that training environments are known to the agent beforehand.
Extending IRM to the distributed setting can be done using game theory (\irmg \cite{ahuja2020invariant}). 
Therein, different agents, each equipped with its own predictor $\bm{w}_r$, cooperate to train an ensemble model: $\bm{w}^{\text{av}} =  \sum_{r \in \mathcal{R}}\frac{D_r}{\sum_{n \in \mathcal{R}}D_n} \bm{w}_r$. 
In contrast to \eqref{eq:IRM} that designs a unique predictor across training environments, the aggregate model $\bm{w}^{\text{av}}$ satisfies:
\begin{equation}
    \label{eq:IRMGames}
    \begin{aligned}
        &\underset{\bm{\phi}, \, \bm{w}^{\text{av}}}{\textrm{minimize}}&& \quad 
        \frac{1}{\lvert\mathcal{R}\rvert} \sum_{r=1}^{\lvert\mathcal{R}\rvert}
        \ell_r\left(f_{\bm{w}^{\text{av}}} \circ f_{\bm{\phi}}\right)\\
        &\textrm{subject to}&& \textstyle
        \bm{w}_r \in \underset{\bm{w}^\prime_r}{\arg\min}\,\;
        \ell_r \bigg( \frac{f_{\bm{w}^\prime_r} + \sum\limits_{n \in \mathcal{R}\backslash\{r\}}f_{\bm{w}_n}}{\lvert\mathcal{R}\rvert} \circ f_{\bm{\phi}} \bigg) \\
        &&& \qquad\qquad\qquad \qquad\qquad\qquad \forall r \in \mathcal{R}.
    \end{aligned}
\end{equation}
The set of constraints in \eqref{eq:IRMGames} represents the Nash Equilibrium of a game where the players/environments $r \in \mathcal{R}$ select actions $\bm{w}_k$ in order to minimize their cost functions $\ell_r\left(f_{\bm{w}^{\text{av}}} \circ f_{\bm{\phi}}\right)$. Since there are no algorithms that guarantee reaching the Nash point of the aforementioned continuous non-zero sum game, Best Response Dynamics (BRD) is used due to its simplicity. It is worth mentioning that \cite{ahuja2020invariant} also showed that the feature extraction map can be fixed to identity $f_{\bm{\phi}} = \mathrm{I}$, and the overall prediction network $f_{\bm{w}} \circ f_{\bm{\phi}}$ will be recovered by the predictor $f_{\bm{w}}$ solving \eqref{eq:IRMGames}. This version of the algorithm is called \firmg, while the one where the extractor and predictor are learned separately is referred to as \virmg.

\irmg is a very suitable fit for FL since it allows agents distributed in different environments to train a collective IRM-based model.
Hence, the authors in \cite{gupta2022fl} proposed minor modifications in order to adapt it to the FL setup. 
First, parallelized BRD is used to mitigate the sequential dependencies and accelerate convergence. 
Moreover, \virmg requires an extra round of optimization for the extraction function parameter $\bm{\phi}$, which delays its convergence. 
Thus, the stochastic gradient descent (SGD) over $\bm{\phi}$ is replaced by a Gradient Descent (GD) update that takes larger steps in the direction of the global optimum. 
In contrast to \cite{gupta2022fl}, which considered highly correlated datasets, the data in our setting shows negligible oscillations in parameter updates under BRD, in which, we do not adopt buffers in training. These buffers can be used by each agent to store the historically played actions of its opponents. Then, an agent responds to a uniform distribution over these past actions. This smoothens the oscillations of BRD caused by the local correlations in the datasets. Finally, the detailed steps of the \vflg algorithm that trains both a representation learner and a predictor are presented in Algorithm \ref{alg:FLGames}.

\section{Simulation Results}\label{section:4}
\subsection{Simulation Settings}
For our experiments, we consider three different environments $\mathcal{R}$. 
In all three environments, the Tx is located at the coordinate $(0, 35, 3)$ and the RIS comprising $N = 10 \times 10$ reflective elements is at $(10, 20, 1)$, with the coordinates given in meters within a Cartesian system. 
The Rx is located in an annular region around the RIS with inner and outer radii $R_{\text{min}} = 1\,$m and $R_{\text{max}} = 5\,$m.
% consisting of receivers that are differently distributed. 
%
The Rician factor $\kappa$ is set to $5$, and the pathloss coefficients are calculated by $\alpha_t = \frac{N d_{\mathrm{x}} d_{\mathrm{y}}}{4 \pi d_t^2}$ and $\alpha_r = \frac{N d_{\mathrm{x}} d_{\mathrm{y}}}{4 \pi d_r^2}$. To simplify the exhaustive search for optimal phases, we assume $\mathcal{C}$ contains two configurations classes, namely $\bm{\theta}_1 = \left[0,0,\dots,0\right]^\trans$ and $\bm{\theta}_2 = \left[0,\pi,0,\pi,\dots,0,\pi\right]^\trans$. 

\begin{figure}
    \centering
    \includegraphics[width=\linewidth]{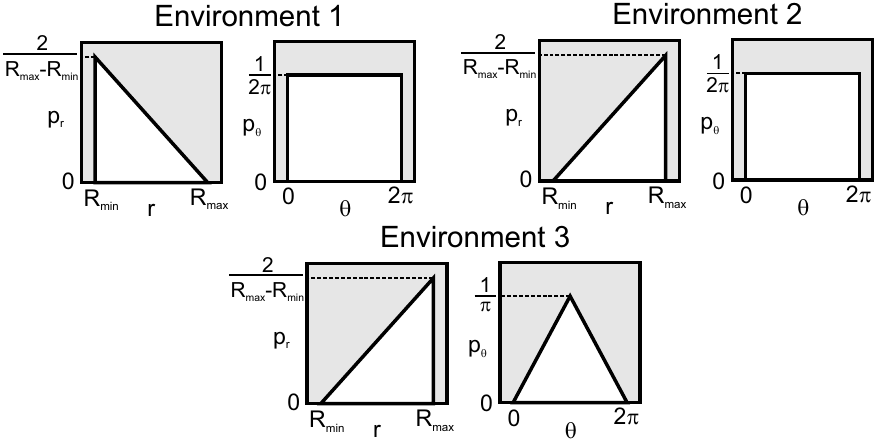}
    \caption{Distributions of the receiver's position (distance $r$ and angle $\theta$ from the RIS) in different environments.}
    \label{fig:prob}
\end{figure}

In environment 1, the RIS elements are distanced by $d_{\mathrm{x}} = d_{\mathrm{y}} = 0.5\lambda$.
Therein, the Rx is uniformly placed around the RIS with a tendency to be deployed closer to the RIS following the distributions illustrated under Environment 1 in Fig. \ref{fig:prob}.
In environment 2, the RIS is characterized by $d_{\mathrm{x}} = d_{\mathrm{y}} = 0.25\lambda$.
In contrast to environment 1, here, the Rx is higher likely to be placed far from the RIS (see the distributions under Environment 2 in Fig. \ref{fig:prob}).
The RIS in environment 3 has $d_{\mathrm{x}} = d_{\mathrm{y}} = 0.4\lambda$.
Therein, the Rx's distance to the RIS is uniform but the angle distribution is concentrated in one direction as illustrated in Fig. \ref{fig:prob} under Environment 3.
Note that the data from environments 1 and 2 are used to compose the training data while environment 3 is used only for testing.

% Fig. \ref{fig:prob} shows the Rx's distance and angle distribution relative to the RIS. In environment 1, the Rx tends to be close to RIS 1 unlike in environment 2, where the Rx's distance to RIS 2 is uniformly distributed in the annulus. In both environments 1 and 2, the Rx's angle is uniformly distributed. We further define environment 3 for testing purposes, where the Rx's distance to the RIS is uniform but the angle distribution is concentrated in one direction. The RIS in environment 3 has $d_{\mathrm{x}} = d_{\mathrm{y}} = 0.4\lambda$.

% The Rician factor $\kappa$ is set to $5$, and the pathloss coefficients are calculated by $\alpha_t = \frac{N d_{\mathrm{x}} d_{\mathrm{y}}}{4 \pi d_t^2}$ and $\alpha_r = \frac{N d_{\mathrm{x}} d_{\mathrm{y}}}{4 \pi d_r^2}$. To simplify the exhaustive search for optimal phases, we assume $\mathcal{C}$ contains two configurations classes, namely $\bm{\theta}_1 = \left[0,0,\dots,0\right]$ and $\bm{\theta}_2 = \left[0,\pi,0,\pi,\dots,0,\pi\right]$. 

\begin{algorithm}
\caption{\flg for RIS}\label{alg:cap}
\renewcommand{\algorithmicrequire}{\textbf{Inputs:}}
\renewcommand{\algorithmicensure}{\textbf{Outputs:}}
\begin{algorithmic}
\Require Set of RISs: $\mathcal{R}$, Datasets: $\mathcal{D}_r$, Learning rates: $\eta_{\bm{w}}, \eta_{\bm{\phi}}$, Number of rounds: $\mathrm{rounds}$, Mini-batch size: $m$
\Ensure $f_{\bm{\phi}}, f_{\bm{w}^{\text{av}}}$\\
\textbf{Server executes:}
\Indent
\State Initialize $\bm{\phi}$ and $\{\bm{w}_r\}_{r\in\mathcal{R}}$
\State Broadcast $\bm{\phi}$ and $\{\bm{w}_r\}_{r\in\mathcal{R}}$ to all agents $r \in \mathcal{R}$
\State Set $\mathrm{round}\gets 1$
\While{$\mathrm{round} \leq \mathrm{rounds}$}
\For{each agent $r \in \mathcal{R}$ \textbf{parallel}}
\State Compute $\nabla\bm{\phi}_r \gets \nabla_{\bm{\phi}} \, \ell_r\left(f_{\bm{w}^{\text{av}}} \circ f_{\bm{\phi}}; \mathcal{D}_r\right)$
\State Communicate $\nabla\bm{\phi}_r$ to the server
\EndFor
\State Update extractor:
\State\hspace{0.8em} $\bm{\phi} \gets \bm{\phi} - \eta_{\bm{\phi}} \sum_{r \in \mathcal{R}}\frac{D_r}{\sum_{n \in \mathcal{R}}D_n}\nabla\bm{\phi}_k$
\State Broadcast $\bm{\phi}$ to all agents $r \in \mathcal{R}$
\For{each agent $r \in \mathcal{R}$ \textbf{parallel}}
\State Sample mini-batch $\mathcal{B}_r$ of size $m$ from $\mathcal{D}_r$
\State Update predictor:
\State\hspace{0.8em} $\bm{w}_r \gets \bm{w}_r - \eta_{\bm{w}}\nabla_{\bm{w}_r} \, \ell_r\left(f_{\bm{w}^{\text{av}}} \circ f_{\bm{\phi}}; \mathcal{B}_r\right)$
\State Communicate $\bm{w}_r$ to the server
\EndFor
\State Average Predictor: $\bm{w}^{\text{av}} \gets \sum_{r \in \mathcal{R}}\frac{D_r}{\sum_{n \in \mathcal{R}}D_n} \bm{w}_k$
\State Broadcast $\{\bm{w}_r\}_{r\in\mathcal{R}}$ to all agents $r \in \mathcal{R}$
\State $\mathrm{round}\gets \mathrm{round} + 1$
\EndWhile
\EndIndent
\end{algorithmic}\label{alg:FLGames}
\end{algorithm}

In this work, we leverage \vflg that exhibits a superior performance than \fflg $\left(\text{where } f_{\bm{\phi}} = \mathrm{I}\right)$.
On the other hand, the authors in \cite{9593252} showed by empirical simulations, due to the fact that RIS channels have strong LoS components, that one can select fixed causal representations based on the channels in \eqref{eq:channelg} and \eqref{eq:channelh}. The causal representations in this case are the AoA and AoD at the RIS, $\left(\varphi_t, \vartheta_t\right)$ and $\left(\varphi_r, \vartheta_r\right)$, and the relative distances RIS-Tx $d_t$ and RIS-Rx $d_r$. This benchmark variant of \flg where $f_{\bm{\phi}}$ is fixed to $\left(\varphi_t, \vartheta_t, \varphi_r, \vartheta_r, d_t, d_r\right)$ is called \fflg in this paper, and is not to be confused with \fflg in \cite{gupta2022fl}, where $f_{\bm{\phi}} = \mathrm{I}$.

\begin{figure}
    \centering
    \includegraphics[height=3.6cm]{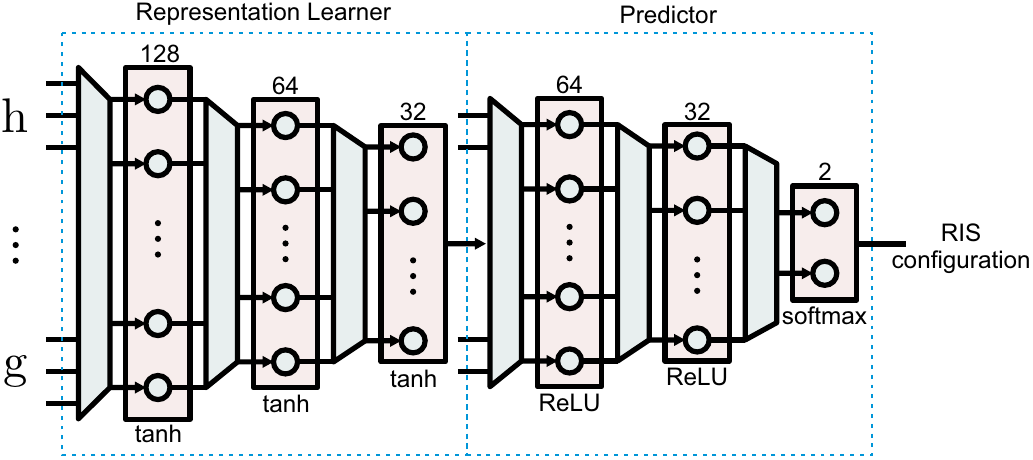}
    \caption{Structure of the neural networks used for the representation learner and the predictor: circles represent activation functions and trapezoids correspond to the weights and biases. The activation function type and the output dimension are shown at the bottom and top of each layer.}
    \label{fig:nn}
\end{figure}

\begin{figure*}
    \begin{subfigure}[t]{0.5\textwidth}%
    \centering\captionsetup{width=0.85\linewidth}%
        \includegraphics[width=\linewidth]{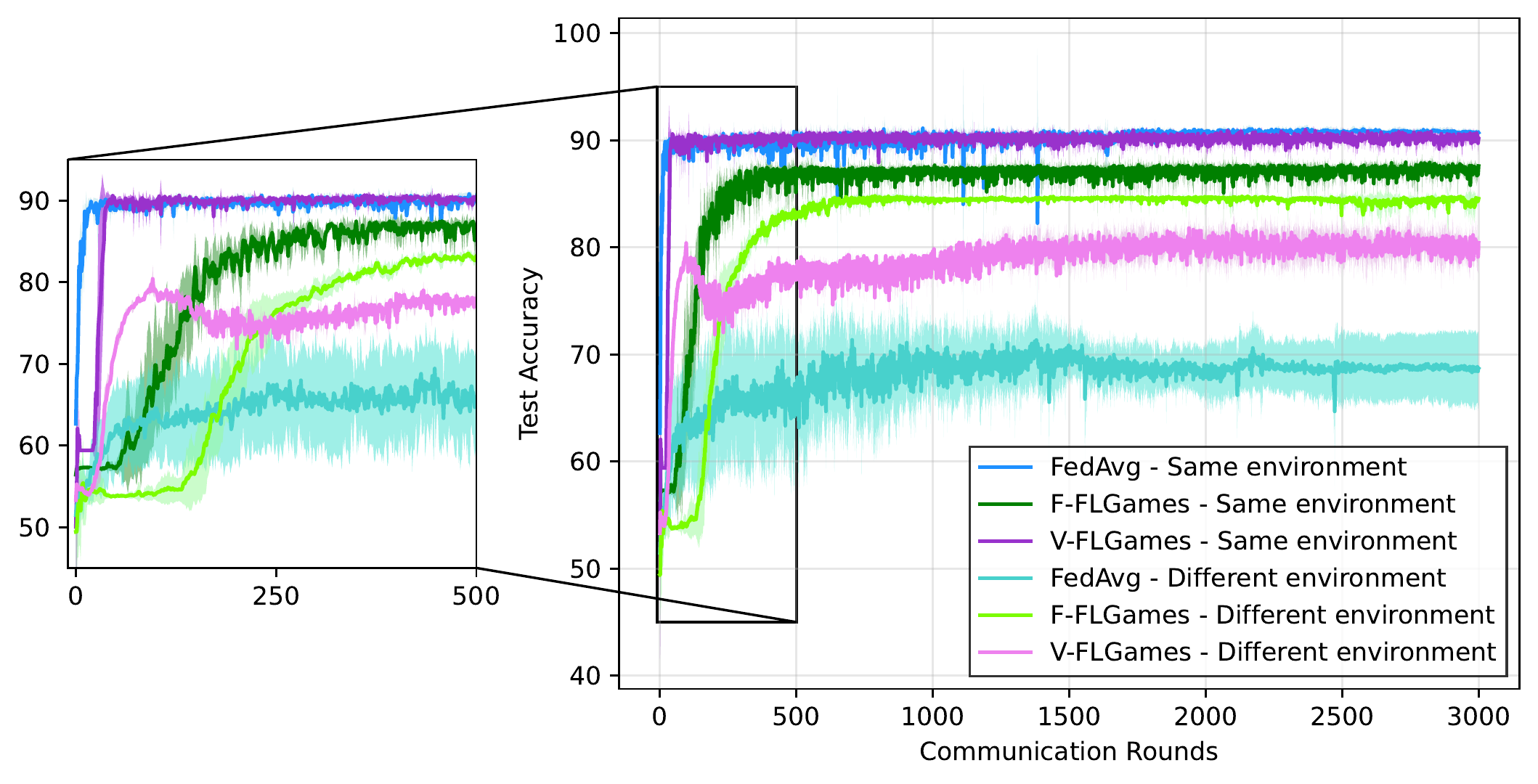}%
        \caption{Test accuracy convergence for all algorithms compared in the same or in a different environment.}
        \label{fig:sim1}
    \end{subfigure}%
    \setcounter{subfigure}{4}
    \hspace{.05\linewidth}%
    \begin{subfigure}[t]{0.342\textwidth}%
    \centering\captionsetup{width=1.2\linewidth}%
        \includegraphics[width=\linewidth]{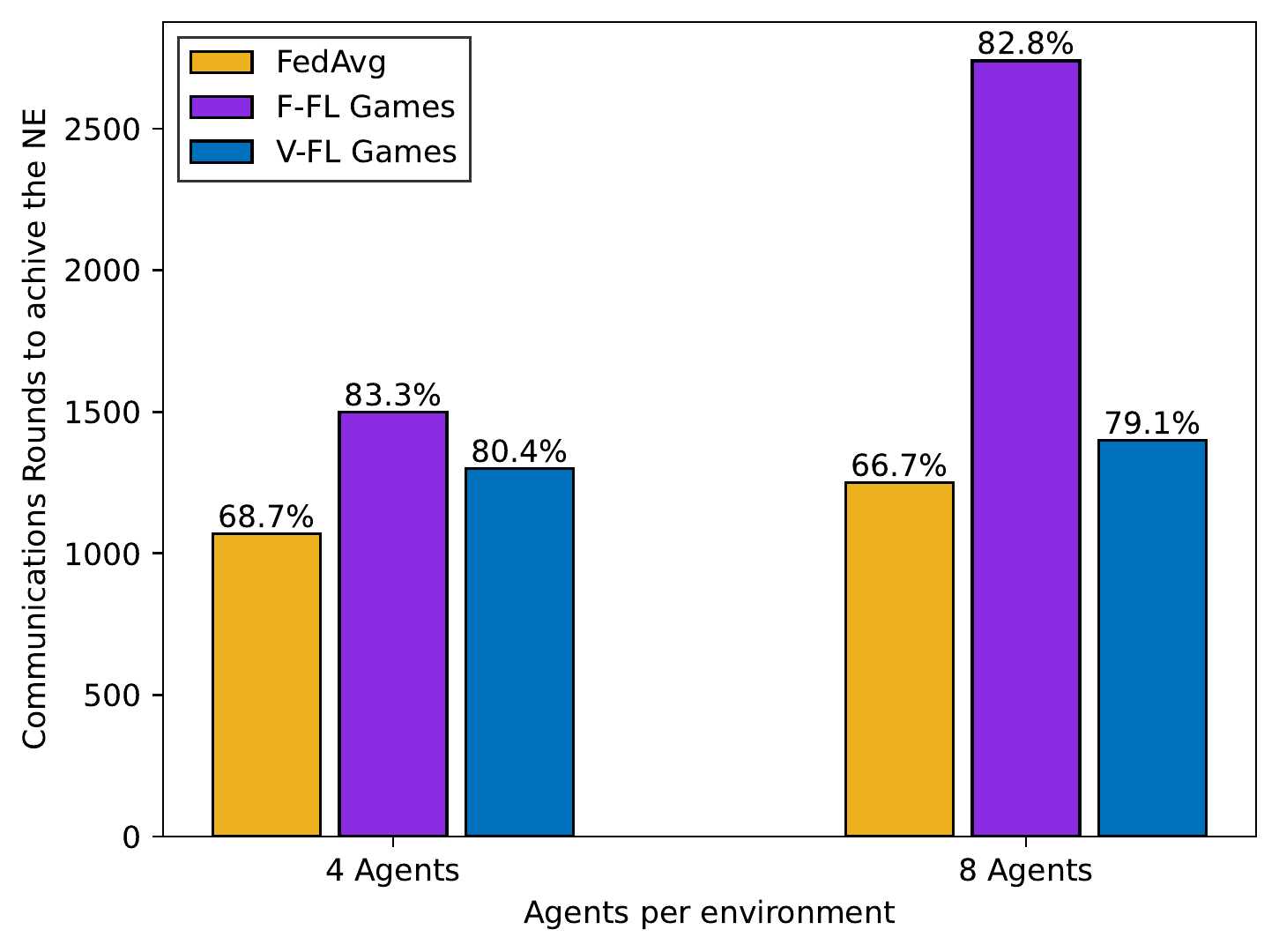}%
        \caption{Communication rounds needed for convergence with multiple agents per environment. The number on top of the bar is the corresponding test accuracy.}
        \label{fig:sim2}
    \end{subfigure}%
    
    \vspace*{\floatsep}
    \setcounter{subfigure}{1}%
    
    \begin{subfigure}{0.32\textwidth}%
    \centering\captionsetup{width=.8\linewidth}%
        \includegraphics[width=0.9\linewidth]{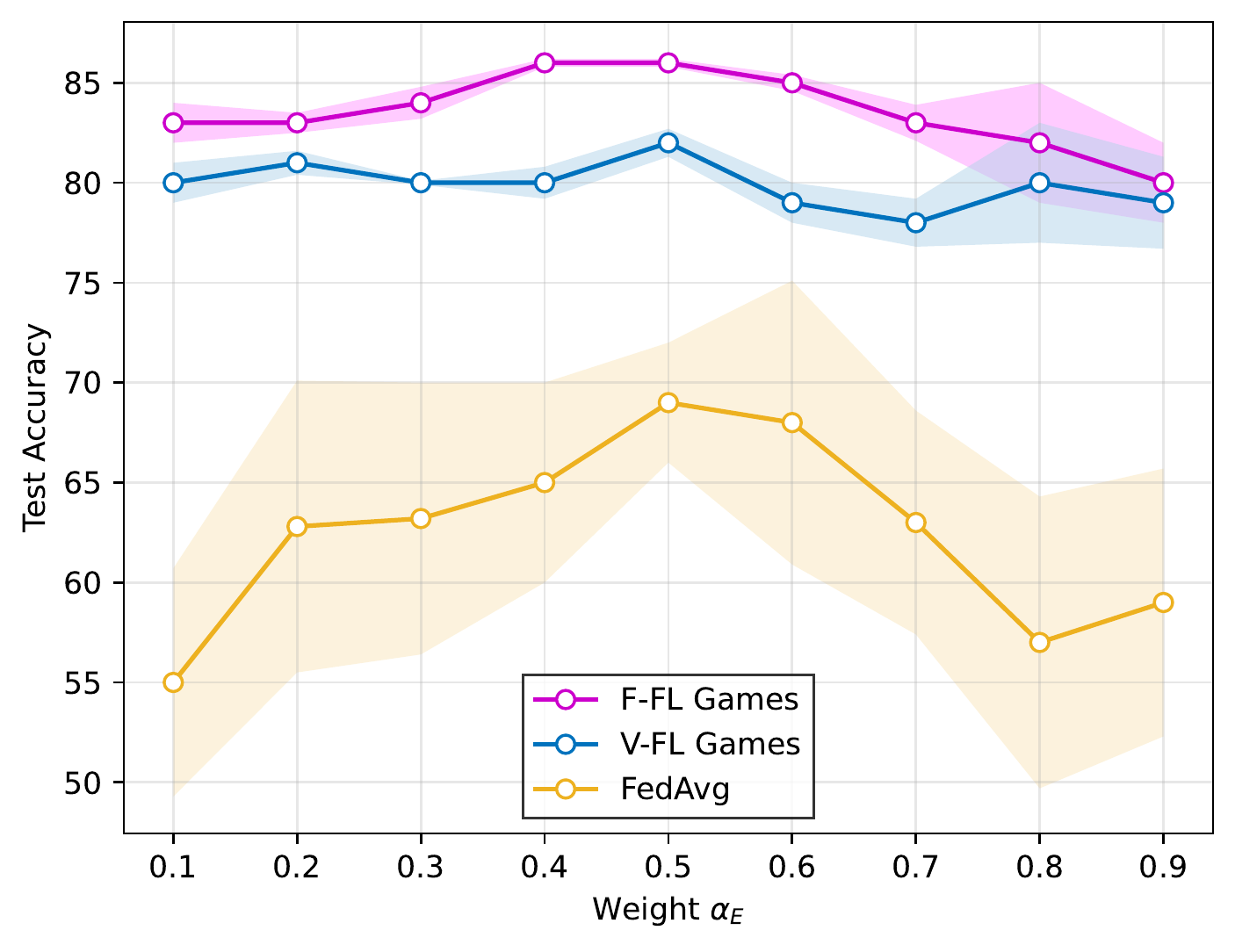}%
        \caption{Impact of the weight $\alpha_e$ on the testing accuracy.}
        \label{fig:sim3}
    \end{subfigure}%
    \begin{subfigure}{0.32\textwidth}%
    \centering\captionsetup{width=.8\linewidth}%
        \includegraphics[width=0.9\linewidth]{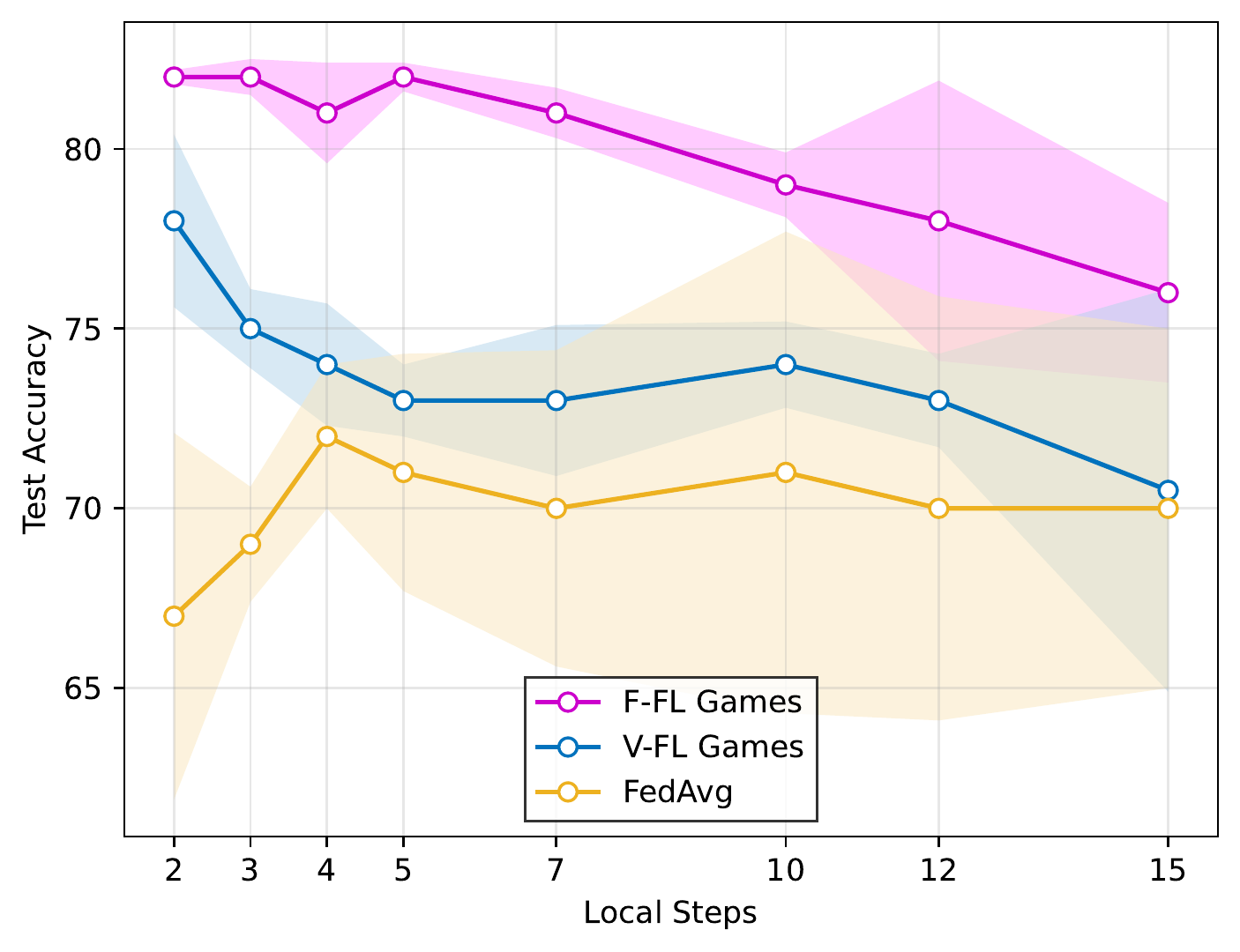}%
        \caption{Impact of number of local iterations on the testing accuracy.}
        \label{fig:sim4}
    \end{subfigure}%
    \begin{subfigure}{0.33\textwidth}%
    \centering\captionsetup{width=\linewidth}%
        \includegraphics[width=0.9\linewidth]{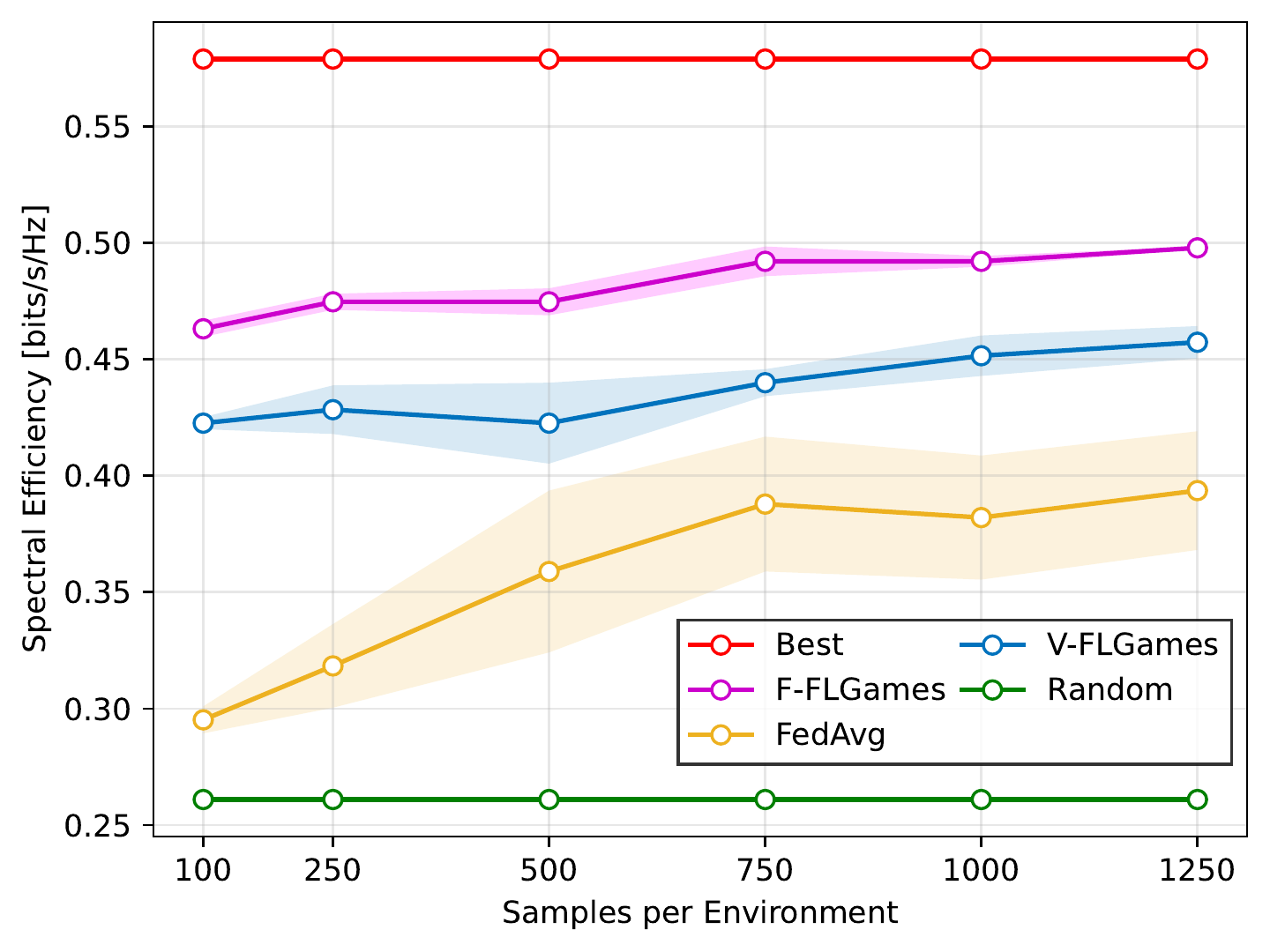}%
        \caption{Impact of the number of samples per environment on the achievable spectral efficiency.}
        \label{fig:sim5}
    \end{subfigure}
\caption{Performance comparison of different algorithms in OoD testing datasets.}
\label{fig:simulations}
\end{figure*}

For training in \fedavg and \vflg, we collect $D_r = 1500$ CSI samples $\bm{x}_j^r = \left(\mathbf{h}_j^r, \mathbf{g}_j^r\right)$ from environments 1 and 2, that are decoupled over real and imaginary parts. 
This data is scaled in such a way that the normalized mean is zero and the normalized variance is one.
We also record the parameters $\left(\varphi_t, \vartheta_t, \varphi_r, \vartheta_r, d_t, d_r\right)$ to train \fflg, which are scaled using a minmax scaler that normalizes the data to the interval $[-1,1]$ by dividing by the absolute maximum. 
Unless stated otherwise, we use $1000$ samples collected from environment 3 for testing. 
The design of the neural networks of the extractor and the predictor of \vflg is based on the multi-layer perceptrons architecture, and is shown in Fig. \ref{fig:nn}.
Note that \fedavg and \fflg only use the predictor part. The considered loss function is the cross-entropy. The mini-batch size used for training the predictor is $m = 32$, and the learning rates are fixed at $\eta_{\bm{\phi}} = 5 \times 10^{-4}$ and $\eta_{\bm{w}} = 2 \times 10^{-3}$.
In the figures, lines correspond to the simulation results that are averaged over five runs while the their respective standard deviations are shown as shaded areas.

\subsection{Discussion}
We first plot the evolution of the testing accuracy of all algorithms in Fig.  \hyperref[fig:sim1]{4(a)}. 
Within the same environment (Environments 1 and 2), \fedavg and \vflg perform similarly with an accuracy of $90\%$, while \fflg gives an accuracy of $87\%$. When testing in a different environment (Environment 3), \fedavg's accuracy drops to $68\%$, highlighting its lack of robustness. In this case, \fflg and \vflg yield slightly lower accuracies of $85\%$ and $80\%$, implying they do not overfit to the statistical correlations in the channels. 
It is also interesting to observe the convergence rate of the different methods. We notice that \vflg converges faster than \fflg, requiring around $500$ communication epochs in both cases.

Fig.  \hyperref[fig:sim3]{4(b)} demonstrates the impact of mixing data from different environments on model training.  
Here, we keep the total number samples to be $D_1 + D_2 = 3000$ and vary the amount of data from the two environments, in which, $\alpha_e = \frac{D_1}{D_2}$ represents the fraction of samples of environment 1 compared to environment 2.
% Fig. \hyperref[fig:sim3]{4(b)} displays the performance of the algorithms for different values of $\alpha_e$. 
We first notice that all algorithms reach their peak performance with balanced datasets, i.e. $\alpha_e = 0.5$. When the datatsets are biased towards one of the environments, \fedavg loses $15\%$ of its performance. On the other hand, \fflg and \vflg maintain a steady accuracy with a slight degradation of about $3$--$4\%$.

Inspired by \fedavg's flexibility in allowing more local computations at each agent before sharing their models, we modify our proposed algorithm to study the effect of the number of local iterations on the model accuracy. 
Insofar, each agent performs a few SGD updates locally prior to model sharing. 
Fig.  \hyperref[fig:sim4]{4(c)} shows the impact of the number of local steps on the testing accuracy. 
It can be noted that the accuracy drops when each RIS performs more local computations when using \flg due to the fact that the models are overfitting to the local datasets.
% Intuitively, by increasing this number of steps, each agent runs over more samples from its local dataset, decreasing the testing accuracy at equilibrium, \textcolor{red}{since the played strategies do not account for the opponents' actions}. 
% Fig. \hyperref[fig:sim4]{4(c)} shows the impact of the number of local steps on the testing accuracy. 
%As expected, the accuracy drops when each RIS performs more local computations when using \flg. 
However, the performance of \fflg is more consistent with the local steps, losing only $2\%$ of accuracy with seven local iterations, while \vflg loses around $8\%$ of accuracy with $15$ local updates.
The reason for this behavior is that by letting each agent run over more samples from its local dataset, the testing accuracy at equilibrium decreases, since the played strategies do not account for the opponents' actions.
On the other hand, the accuracy of \fedavg slightly increases with more local iterations, but still performs poorly.

The effect of the dataset size $D_r$ per agent on the achievable spectral efficiency is illustrated in Fig.  \hyperref[fig:sim5]{4(d)}.
% We now study the effect of the number of samples $D_r$ per agent. 
Note that all agents use an equal amount of samples, i.e., $\alpha_e = 0.5$ is held. 
For the comparison, we additionally present the optimal rate given by the best configurations (indicated by Best) and the rates given by random phase decision making (indicated by Random).
All algorithms reach their best performance with the highest number of samples $D_r=1250$ with about $21\%$, $14\%$ and $32\%$ losses compared to Best rates in \vflg, \fflg, and \fedavg, respectively.
The advantage of learning invariant causal representations with minimal amount of data is highlighted when $D_r\leq 750$. 
\fedavg looses its performance rapidly.
Even with $100$ samples per environment, \flg algorithms lose only $6\%$ of their performance, while \fedavg incurs more than $15\%$ of its accuracy. \fedavg requires around $750$ samples per agent to reach its best performance, that is more than $10\%$ less than that given by \vflg, underscoring its weakness in OoD settings. Additionally, with $1250$ samples per environment, the error variance of \vflg and \fflg is $73\%$ and $98\%$ less than that of \fedavg.

Finally, we vary the number of agents per environment as shown in Fig. \hyperref[fig:sim2]{4(e)}. For this experiment, $1500$ samples from each environment are shared among all agents, so more agents having less data are involved. The achieved testing accuracies of the \flg algorithms are still superior than the \fedavg benchmark. Surprisingly, doubling the number of collaborating RISs from $8$ to $16$ induces an $82\%$ increase in the number of training epochs for convergence in \fflg. The same does not hold for \vflg that suffers from an $8\%$ increase, while losing $3$-$4\%$ in accuracy compared to \fflg. This implies that, with more agents owning fewer data, the training of a causal extractor and a predictor converges faster than training of only a predictor.

\section{Conclusions}\label{section:5}
This paper proposes a distributed phase configuration control for RIS-assisted communication systems. The rate maximization problem is formulated using federated IRM as opposed to a heterogeneity-unaware ERM approach. Our novel robust RIS phase-shifts controller leverages the underlying causal representations of the data that are invariant over different environments. A neural network based  feature extractor first uncovers the causal structure of the CSI data, then feeds it to another neural network based configuration predictor. Both neural networks are trained in a distributed supervised learning fashion, and the results are compared with the environment-unaware \fedavg and an IRM-based predictor. The numerical results show that a phase predictor trained with the geometric properties of the environments demonstrated a better performance than a representation learner followed by a predictor. Moreover, the extractor-predictor network exhibits faster training convergence when using more RISs.
The extensions for multiple users and multiple antennas at Tx and Rx are left for future works.

\bibliographystyle{IEEEtran}
\bibliography{references}

\end{document}